\documentclass[]{interact}

\usepackage{epstopdf}
\usepackage{subfigure}
\usepackage{tabularx}
\usepackage{graphicx}
\usepackage{amsmath}
\usepackage{amssymb}
\usepackage{booktabs}
\usepackage{placeins} 
\usepackage{float} 
\usepackage{enumitem}
\usepackage{tabularx} 
\usepackage{longtable} 
\usepackage{array} 
\usepackage{ragged2e} 
\usepackage{booktabs} 
\usepackage{placeins}
\usepackage{afterpage} 
\usepackage{caption} 
\usepackage{booktabs} 
\usepackage{adjustbox} 

\begin{document}

\title{3D Wavelet Convolutions with Extended Receptive Fields for Hyperspectral Image Classification}

\author{
\name{Guandong Li\textsuperscript{a}\thanks{CONTACT Guandong Li. Email: leeguandon@gmail.com} and Mengxia Ye\textsuperscript{b}}
\affil{\textsuperscript{a}iFLYTEK, Shushan, Hefei, Anhui, China; \textsuperscript{b}Aegon THTF,Qinghuai,Nanjing,Jiangsu,China}
}

\maketitle

\begin{abstract}
Deep neural networks face numerous challenges in hyperspectral image classification, including high-dimensional data, sparse ground object distributions, and spectral redundancy, which often lead to classification overfitting and limited generalization capability. To better adapt to ground object distributions while expanding receptive fields without introducing excessive parameters and skipping redundant information, this paper proposes WCNet, an improved 3D-DenseNet model integrated with wavelet transforms. We introduce wavelet transforms to effectively extend convolutional receptive fields and guide CNNs to better respond to low frequencies through cascading, termed wavelet convolution. Each convolution focuses on different frequency bands of the input signal with gradually increasing effective ranges. This process enables greater emphasis on low-frequency components while adding only a small number of trainable parameters. This dynamic approach allows the model to flexibly focus on critical spatial structures when processing different regions, rather than relying on fixed receptive fields of single static kernels. The Wavelet Conv module enhances model representation capability by expanding receptive fields through 3D wavelet transforms without increasing network depth or width. Experimental results demonstrate superior performance on the IN, UP, and KSC datasets, outperforming mainstream hyperspectral image classification methods.
\end{abstract}


\section{Introduction}

Hyperspectral remote sensing images (HSI) play a crucial role in spatial information applications due to their unique narrow-band imaging characteristics. Imaging equipment synchronously records both spectral and spatial position information, integrating them into a three-dimensional data structure containing two-dimensional space and one-dimensional spectrum. As an important application of remote sensing technology, ground object classification demonstrates broad value in ecological assessment, transportation planning, agricultural monitoring, land management, and geological surveys \cite{chang2003hyperspectral}. However, HSI classification faces several challenges. First, hyperspectral data typically has high-dimensional characteristics, with each pixel containing hundreds or even thousands of bands, leading to data redundancy and computational complexity while potentially causing the "curse of dimensionality," making classification models prone to overfitting under sparse sample conditions. Second, sparse ground object distribution means training samples are often limited, especially for certain rare categories where annotation acquisition is costly and imbalanced, further restricting model generalization. Additionally, HSIs are often affected by noise, atmospheric interference, and mixed pixels, reducing signal-to-noise ratio and increasing feature extraction difficulty for sparse objects. Finally, ground object sparsity may lead to insufficient spatial context information, causing classification errors.

Deep learning methods for HSI classification \cite{li2025spatial,li2018scene,li2023dgcnet,li2025efficient,li2019doubleconvpool,li2025spatialgeometry} have made substantial progress. In \cite{lee2017going} and \cite{zhao2016spectral}, Principal Component Analysis (PCA) was first applied to reduce dimensionality before extracting spatial information using 2D CNN. Methods like 2D-CNN \cite{makantasis2015deep, chen2016deep} require separate extraction of spatial and spectral features, failing to fully utilize joint spatial-spectral information while needing complex preprocessing. \cite{wang2018fast} proposed a Fast Dense Spectral-Spatial Convolutional Network (FDSSC) using serially connected 1D-CNN and 3D-CNN dense blocks. FSKNet \cite{li2022faster} introduced a 3D-to-2D module and selective kernel mechanism, while 3D-SE-DenseNet \cite{li2020hyperspectral} incorporated the SE mechanism into 3D-CNN to correlate feature maps across channels, activating effective information while suppressing ineffective features. DGCNet \cite{li2023dgcnet} designed dynamic grouped convolution (DGC) on 3D kernels, where each group includes small feature selectors to dynamically determine input channel connections based on activations, enabling CNNs to learn rich feature representations. DHSNet \cite{liu2025dual} proposed a Center Feature Attention-Aware Convolution (CFAAC) module that focuses on cross-scene invariant features, enhancing generalization capability. Therefore, 3D-CNN has become a primary approach for HSI classification.

To leverage both CNN and Transformer advantages, many studies combine them to utilize local and global features. \cite{sun2022spectral} proposed a Spectral-Spatial Feature Tokenization Transformer (SSFTT) using 3D/2D convolutional layers for shallow features and Gaussian-weighted tokens in transformer encoders for high-level semantics. Some Transformer-based methods \cite{hong2021spectralformer} employ grouped spectral embedding and transformer encoders, but these treat spectral bands or spatial patches as tokens, causing significant redundant computations. Given HSI's inherent redundancy, these methods often underperform 3D-CNN approaches while requiring greater complexity.

3D-CNN can simultaneously sample spatial and spectral dimensions while preserving 2D convolution's spatial feature extraction capability. However, the 3D-CNN paradigm has two major limitations: (1) When simultaneously extracting spatial and spectral features, 3D convolution may incorporate irrelevant or inefficient spatial-spectral combinations into calculations. For instance, certain spatial features may be prominent in specific bands while appearing as noise or irrelevant information in others, yet 3D convolution still forcibly fuses these low-value features, with computational and dimensional redundancy easily triggering overfitting risks that further limit model generalization. (2) The receptive fields of 3D CNNs are inferior to Transformer architectures, and under the sparse ground object characteristics of hyperspectral data, this leads to loss of contextual spatial information and reduced classification accuracy. However, Transformer-like methods also fail to adequately address information redundancy issues, causing increased computation while excessive fitting of redundant information further limits classification capability. Current methods like DFAN \cite{zhang2020deep}, MSDN \cite{zhang2019multi}, 3D-DenseNet \cite{zhang2019three}, and 3D-SE-DenseNet use dense connections that directly link each layer to all preceding layers, enabling feature reuse but introducing redundancy when later layers don't need early features. Therefore, more efficient enhancement of 3D kernel representation while expanding receptive fields to capture sparse objects, and further optimization of redundant information filtering and skipping in dense connections have become directions for hyperspectral classification.

Addressing these issues, \cite{ding2022scaling} used large selective kernels to adjust receptive fields, while \cite{liu2022more} conducted in-depth analysis of Transformer characteristics, treating HSI as 1D sequences that ignore spatial attributes and lack spatial-spectral dependency extraction, employing large kernel spectral networks to expand receptive fields. However, these methods, whether based on Transformers or 3D CNNs, inevitably introduce substantial redundant computation while expanding receptive fields. \cite{gavrikov2024can} revealed an interesting property: using larger kernels makes CNNs more shape-biased, meaning their ability to capture low frequencies in images improves. This finding is somewhat surprising since convolutional layers typically tend to respond to high frequencies in inputs \cite{geirhos2018imagenet,ding2022scaling}. Unlike attention heads that excel at capturing low frequencies, we consider using signal processing tools with global receptive fields to expand convolutional receptive fields without excessively introducing parameters that would make networks redundant. We employ wavelet transforms, a mature tool from time-frequency analysis, to effectively extend convolutional receptive fields and guide CNNs to better respond to low frequencies through cascading. This makes spatial operations (e.g., convolution) in the wavelet domain more meaningful. This paper designs an extended-receptive-field 3D wavelet convolution for joint spatial-spectral hyperspectral image classification (WCNet), proposing Wavelet Convolution that employs cascaded WT decomposition and a series of small-kernel convolutions, with each convolution focusing on different frequency bands of the input signal and effective ranges gradually increasing. This process emphasizes low-frequency components while adding only a small number of trainable parameters. For a k×k effective range, our trainable parameters grow logarithmically with k. Compared to traditional 3D-CNN, this method effectively expands convolutional kernel receptive fields through wavelet convolution without increasing network depth or width, enhancing spatial feature representation across large spaces, particularly for unevenly distributed targets or complex spatial patterns in hyperspectral images, enabling more efficient local feature extraction. Second, the WTConv layer is constructed to better capture low frequencies than standard convolution. Repeated WT decomposition of input low frequencies emphasizes these components and increases layer response. This discussion complements analyses showing convolutional layers respond to high frequencies in inputs. By using compact kernels on multi-frequency inputs, WTConv layers place additional parameters where most needed. Wavelet convolution can reduce aggregation of redundant spectral information, avoiding the uniform sampling and dimensionality reduction treatment of all spectral dimensions in traditional 3D-CNN, effectively alleviating parameter redundancy and computational complexity caused by high spectral dimensionality.

The main contributions are:

1. We propose an extended-receptive-field wavelet convolutional network that improves efficient 3D-DenseNet for joint spatial-spectral HSI classification, using Wavelet Conv to expand receptive fields, address spatial sparsity and information redundancy-induced overfitting risks in hyperspectral data, and enhance network generalization. Combining dense connections with Wavelet Conv, where dense connections facilitate feature reuse in networks, we design a composite 3D-DenseNet model achieving excellent accuracy on both IN and UP datasets.

2. We introduce a wavelet transform-based Wavelet Conv mechanism in 3D-CNN by modifying standard 3DConv with cascaded WT decomposition and a series of small-kernel convolutions, each focusing on different input signal frequency bands with gradually increasing effective ranges. This process emphasizes low-frequency components while adding few trainable parameters, achieving more efficient feature extraction.

3. WCNet is more concise than networks combining various DL mechanisms, without complex connections or serialization, requiring less computation while enhancing model representation through extended-receptive-field wavelet convolution without increasing network depth or width.

\section{3D Wavelet CNN with Extended Receptive Fields}

\subsection{Wavelet Transforms and Large Kernel Convolutions}

Wavelet Transforms (WT) \cite{daubechies1992ten}, a powerful signal processing and analysis tool, have been widely used since the 1980s. Following their success in classical settings, WT has recently been incorporated into neural network architectures for various tasks. \cite{finder2024wavelet} introduced 2D wavelet transforms into classification networks, designing WTConv as a direct replacement for depthwise separable convolution that can be used in any given CNN architecture without additional modifications. By integrating it into ConvNeXt \cite{liu2022convnet} for image classification, they demonstrated its practicality in fundamental vision tasks. Learned Image Compression (LIC) primarily reduces spatial redundancy through autoencoder networks and entropy coding but hasn't completely eliminated frequency-domain correlations through linear transforms like DCT or wavelet transforms - cornerstones of traditional methods. To address this key limitation, \cite{fu2024weconvene} proposed a surprisingly simple yet effective framework incorporating Discrete Wavelet Transform (DWT) into LIC's convolutional layers and entropy coding.

Meanwhile, introducing transformers into vision tasks \cite{liu2021swin} - with their non-local self-attention layers - typically yields better results than local mixing convolutions, reigniting interest in exploring larger CNN kernel sizes. Particularly, \cite{liu2022convnet} revisited the popular ResNet architecture, including empirical comparisons of different kernel sizes, concluding performance saturates at 7×7 kernel sizes. \cite{trockman2022patches} attempted to mimic ViT architecture using only convolutions, demonstrating impressive results by replacing attention (or "mixer") components with 9×9 convolutions. \cite{ding2022scaling} suggested that simply increasing kernel size could harm convolution's local nature, proposing parallel use of small and large kernels then summing their outputs. Using this technique, they successfully trained CNNs with kernel sizes up to 31×31. \cite{liu2022more} increased kernel size to 51×51 by decomposing it into parallel 51×5 and 5×51 kernels, additionally introducing sparsity and network width expansion. However, the idea of more channels (with sparsity) is orthogonal to increasing kernel size. WCNet introduces wavelet transforms into convolution, then more cleverly sums outputs of various frequency components from inputs through the proposed layer to capture multiple receptive fields, thereby expanding receptive fields to achieve the same advantages as transformer mechanisms.

\subsection{Wavelet Convolution with Extended Receptive Fields}

Hyperspectral imagery suffers from limited sample availability and exhibits sparse ground object characteristics with uneven spatial distribution, while containing substantial high-dimensional redundant information in the spectral domain. Although 3D-CNN architectures can leverage joint spatial-spectral information, how to achieve more effective deep extraction of spatial-spectral features remains a critical research challenge. As the core component of convolutional neural networks, convolutional kernels are typically regarded as information aggregators that combine spatial information and feature-dimensional data within local receptive fields. Composed of multiple convolutional layers, nonlinear layers, and downsampling layers, CNNs can capture image features from global receptive fields for comprehensive image representation. However, training high-performance networks remains challenging, with numerous studies focusing on spatial dimension improvements. For instance, residual structures enhance deep feature extraction by fusing outputs from different blocks, while DenseNet improves feature reuse through dense connections.

In feature extraction, 3D-CNNs process both spatial and spectral information of hyperspectral imagery through convolutional operations. However, their kernels often contain numerous redundant weights that contribute minimally to the final output. This redundancy is particularly prominent in joint spatial-spectral feature extraction: From the spatial perspective, the sparse and uneven distribution of ground objects may lead convolutional kernels to capture numerous irrelevant or low-information regions within local receptive fields. Spectrally, hyperspectral data typically contains hundreds of bands with high inter-band correlation and redundancy, making it difficult for convolutional kernels to effectively focus on discriminative features along the spectral axis. The spectral redundancy causes many parameters to merely serve as "fillers" in high-dimensional processing, failing to fully exploit deep patterns in joint spatial-spectral information. This weight redundancy not only increases computational complexity but may also impair the model's capability to represent sparse objects and complex spectral signatures, ultimately limiting 3D-CNN performance in hyperspectral image analysis.

This paper introduces wavelet convolution with extended receptive fields for hyperspectral image processing. First, each WT level increases the layer's receptive field size by adding only a small number of trainable parameters. That is, $l$-level cascaded frequency decomposition of WT plus fixed-size kernels $k$ at each level allows parameter count to grow linearly with level count ($l \cdot 4 \cdot c \cdot k^2$) while receptive fields grow exponentially ($2^l \cdot k$). Second, the WTConv layer is constructed to better capture low frequencies than standard convolution. Repeated WT decomposition of input low frequencies emphasizes these components and increases layer response. This discussion complements analyses showing convolutional layers respond to high frequencies in inputs. By using compact kernels on multi-frequency inputs, WTConv layers place additional parameters where most needed. This ultimately achieves effective spatial-spectral feature representation extraction, particularly valuable for sparse ground objects in hyperspectral data where limited samples (small-sample scenario) lead to significant feature map variations across different kernels. By combining wavelet convolution with 3D-DenseNet's hierarchical characteristics, we enable more effective feature extraction.

\subsubsection{Wavelet Transform as Convolution}
In this work, we employ Haar Wavelet Transform (Haar WT) for its efficiency and directness \cite{finder2022wavelet}. Given an image $X$, a single-level Haar WT along one spatial dimension (width or height) is implemented by depthwise convolution with kernels $[1,1]/2$ and $[1,-1]/2$, followed by application of a standard $2\times$ downsampling operator. To perform 2D Haar WT, we combine this operation along both dimensions, resulting in a depthwise convolution with stride 2 using the following set of four filters:

\begin{equation}
f_{LL}=\frac{1}{2}\begin{bmatrix}1 & 1\\1 & 1\end{bmatrix}, \quad
f_{LH}=\frac{1}{2}\begin{bmatrix}1 & -1\\1 & -1\end{bmatrix}, \quad
f_{HL}=\frac{1}{2}\begin{bmatrix}1 & 1\\-1 & -1\end{bmatrix}, \quad
f_{HH}=\frac{1}{2}\begin{bmatrix}1 & -1\\-1 & 1\end{bmatrix}
\end{equation}

Note that $f_{LL}$ is a low-pass filter while $f_{LH}$, $f_{HL}$, $f_{HH}$ form a set of high-pass filters. For each input channel, the convolution output is:

\begin{equation}
[X_{LL}, X_{LH}, X_{HL}, X_{HH}] = \text{Conv}([f_{LL}, f_{LH}, f_{HL}, f_{HH}], X)
\end{equation}

yielding four channels, each with half the resolution of $X$ (along each spatial dimension). $X_{LL}$ represents $X$'s low-frequency component, while $X_{LH}$, $X_{HL}$, $X_{HH}$ represent its horizontal, vertical, and diagonal high-frequency components respectively.

Since the kernels in Equation (1) form an orthogonal basis, the inverse wavelet transform (IWT) can be obtained via transposed convolution:

\begin{equation}
X = \text{Conv-transposed}([f_{LL}, f_{LH}, f_{HL}, f_{HH}], [X_{LL}, X_{LH}, X_{HL}, X_{HH}])
\end{equation}

Cascaded wavelet decomposition is achieved by recursively decomposing the low-frequency component:

\begin{equation}
X_{LL}^{(i)}, X_{LH}^{(i)}, X_{HL}^{(i)}, X_{HH}^{(i)} = \text{WT}(X_{LL}^{(i-1)})
\end{equation}

where $X_{LL}^{(0)}=X$ and $i$ is the current level. This yields improved frequency resolution while maintaining spatial localization.

\begin{figure}[h]
\centering
\includegraphics[width=0.9\linewidth]{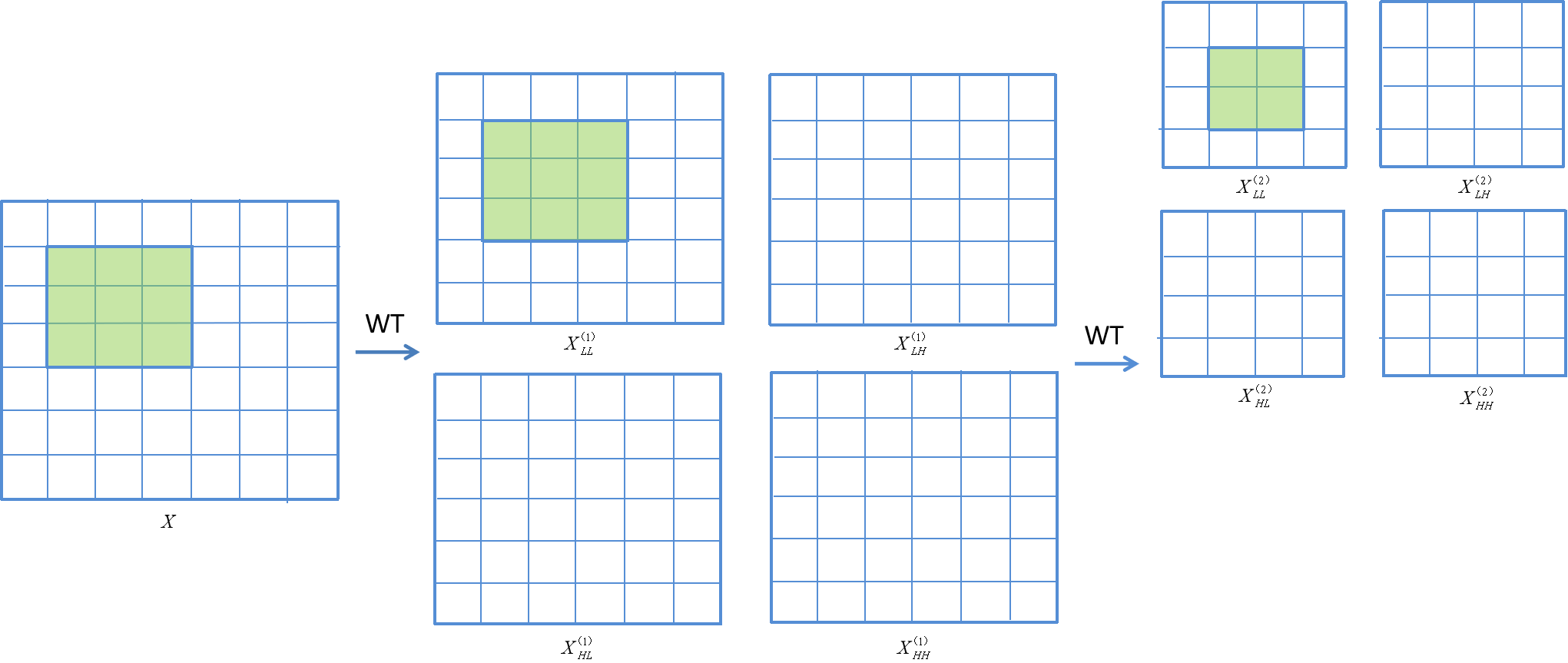}
\caption{Performing convolution in the wavelet domain results in a larger receptive field}
\label{fig:wavelet_receptive}
\end{figure}

\subsubsection{Wavelet Convolution}
As mentioned, increasing convolutional kernel size causes quadratic growth in parameter count (hence degrees of freedom). To mitigate this, we propose the following approach: First, use wavelet transform (WT) to filter and downsample input's low- and high-frequency content. Then, perform small-kernel depthwise convolution on different frequency maps before reconstructing output using inverse wavelet transform (IWT). In other words, the process consists of:

\begin{equation}
Y = \text{IWT}(\text{Conv}(W, \text{WT}(X)))
\end{equation}

where $X$ is the input tensor, $W$ is a weight tensor of depthwise separable convolution kernels with size $k \times k$ and input channels $4 \times$ those of $X$. This operation not only separates convolution between frequency components but also allows smaller kernels to operate on larger regions of original input, effectively increasing its receptive field relative to input.

We adopt this 1-level combined operation and further enhance it using the same cascading principle from Equation (4). The process consists of:

\begin{equation}
X_{LL}^{(i)}, X_{H}^{(i)} = \text{WT}(X_{LL}^{(i-1)})
\end{equation}

\begin{equation}
Y_{LL}^{(i)}, Y_{H}^{(i)} = \text{Conv}(W^{(i)}, (X_{LL}^{(i)}, X_{H}^{(i)}))
\end{equation}

where $X_{LL}^{(0)}$ is the layer's input, and $X_{H}^{(i)}$ represents all three high-frequency maps at the $i$-th level as described in \S3.1.

To combine outputs from different frequencies, we utilize the fact that wavelet transform and its inverse are linear operations, i.e., $\text{IWT}(X+Y) = \text{IWT}(X) + \text{IWT}(Y)$. Thus, performing:

\begin{equation}
Z^{(i)} = \text{IWT}(Y_{LL}^{(i)} + Z^{(i+1)}, Y_{H}^{(i)})
\end{equation}

yields summation of convolutions from different levels, where $Z^{(i)}$ is the accumulated output starting from level $i$, meaning outputs from two differently-sized convolutions are summed as output. We found channel-wise scaling to weigh each frequency component's contribution sufficient.

\begin{figure}[h]
\centering
\includegraphics[width=0.9\linewidth]{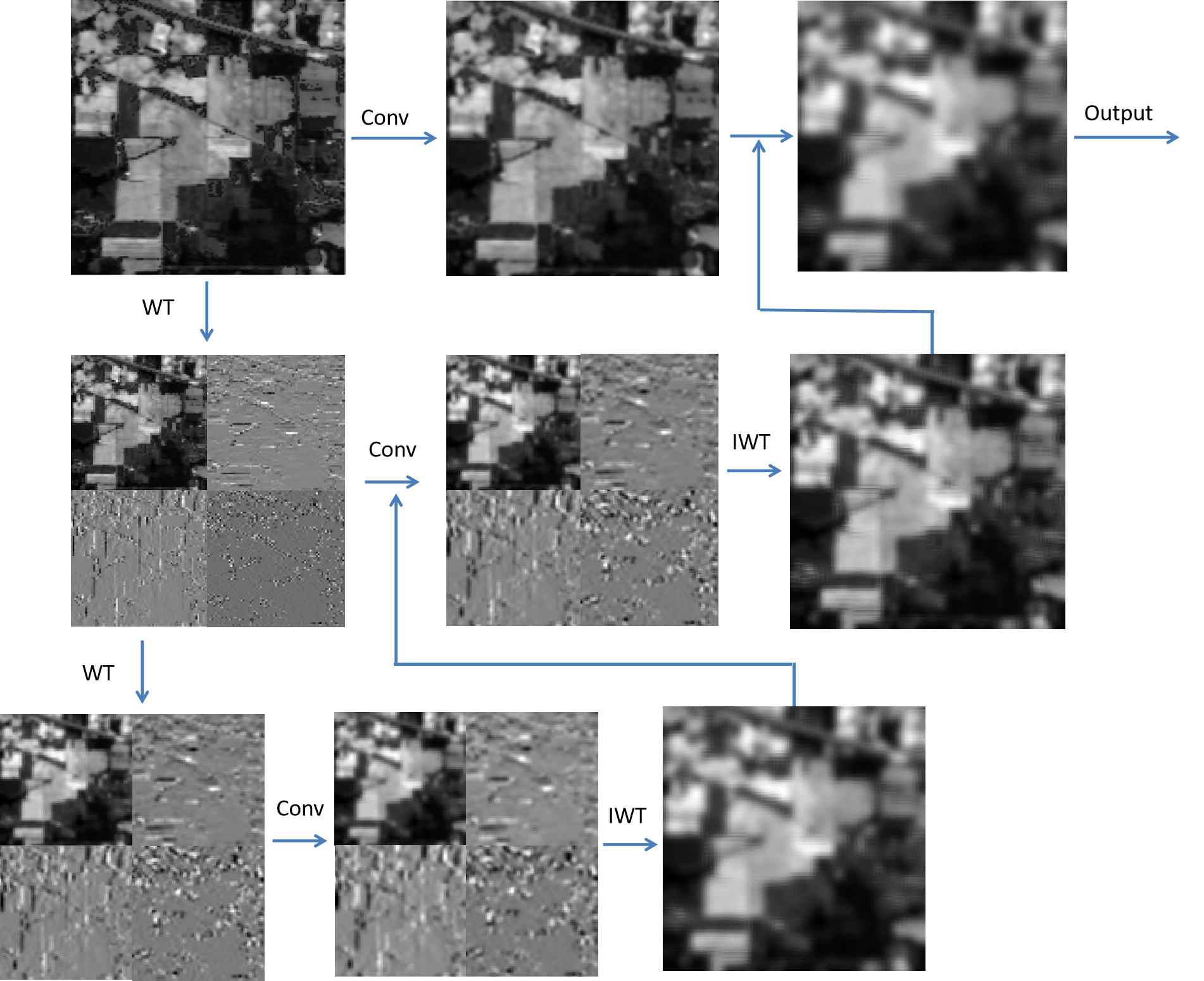}
\caption{The workflow of WTCONV operation}
\label{fig:tconv_workflow}
\end{figure}

Incorporating WTConv into specific CNNs offers two main technical advantages. First, each WT level increases the layer's receptive field size by adding only a small number of trainable parameters. That is, $l$-level cascaded frequency decomposition of WT plus fixed-size kernels $k$ at each level allows parameter count to grow linearly with level count ($l \cdot 4 \cdot c \cdot k^2$) while receptive fields grow exponentially ($2^l \cdot k$).

The second benefit is that the WTConv layer is constructed to better capture low frequencies than standard convolution. Repeated WT decomposition of input low frequencies emphasizes these components and increases layer response. This discussion complements analyses showing convolutional layers respond to high frequencies in inputs. By using compact kernels on multi-frequency inputs, WTConv layers place additional parameters where most needed.

\subsection{Computational Cost}
The computational cost of depthwise convolution (measured in floating-point operations, FLOPs) is:
\begin{equation}
C \cdot K_W \cdot K_H \cdot N_W \cdot N_H \cdot \frac{1}{S_W} \cdot \frac{1}{S_H},
\end{equation}
where $C$ is the number of input channels, $(N_W, N_H)$ are the spatial dimensions of the input, $(K_W, K_H)$ is the kernel size, and $(S_W, S_H)$ are the strides in each dimension. For example, consider a single-channel input with spatial dimensions $512 \times 512$. Using a $7 \times 7$ kernel for convolution results in $12.8$M FLOPs, while a $31 \times 31$ kernel would require $252$M FLOPs.For the WTConv operation set, each wavelet-domain convolution operates on halved spatial dimensions, though with four times the channels of the original input. This leads to a FLOP count of:
\begin{equation}
C \cdot K_W \cdot K_H \cdot \left(N_W \cdot N_H + \sum_{i=1}^{\ell} \frac{4 \cdot N_W \cdot N_H}{2^i \cdot 2^i}\right),
\end{equation}
where $\ell$ is the number of wavelet transform (WT) levels. Taking the previous input size $512 \times 512$ as an example, performing a 3-level WTConv with kernel size $5 \times 5$ (covering a receptive field of $40 \times 40 = (5 \cdot 2^3) \times (5 \cdot 2^3)$) results in $15.1$M FLOPs. Additionally, we must account for the computational cost of the wavelet transform itself. We note that when using the Haar basis, the wavelet transform can be implemented very efficiently. However, if implemented naively using standard convolution operations, the FLOP count for WT would be:
\begin{equation}
4C \cdot \sum_{i=0}^{\ell-1} \frac{N_W \cdot N_H}{2^i \cdot 2^i},
\end{equation}
since there are four $2 \times 2$ kernels with stride 2 in each spatial dimension, operating on each input channel. Similarly, the IWT has the same FLOP count as WT. Continuing this example, this adds $2.8$M FLOPs for 3-level WT and IWT, totaling $17.9$M FLOPs - still significantly more efficient than standard depthwise separable convolution with similar receptive fields.

\begin{figure}[h]
\centering
\includegraphics[width=0.9\linewidth]{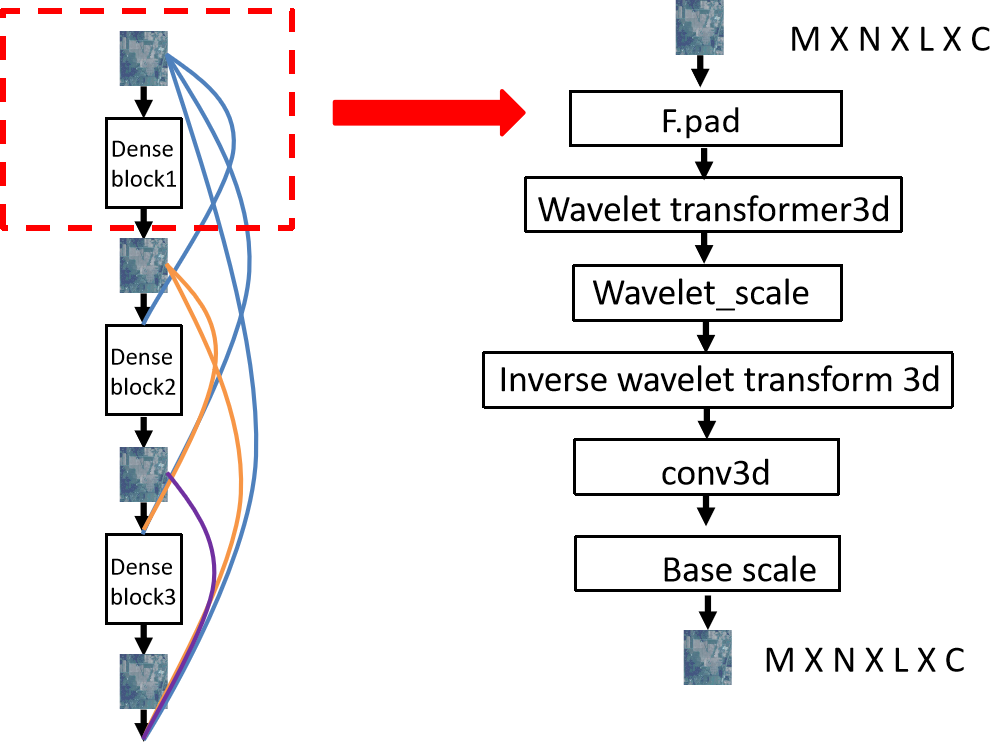}
\caption{Wavelet convolution in 3D-DenseNet's dense block}
\label{fig:denseblock_wavelet}
\end{figure}

\subsection{3D-CNN Hyperspectral Image Feature Extraction Framework and Model Implementation}
We made two modifications to the original 3D-DenseNet, aiming to further simplify the architecture and improve its computational efficiency.

\subsubsection{Exponentially Increasing Growth Rate}
The original DenseNet design adds $k$ new feature maps at each layer, where $k$ is a constant called the growth rate. As shown in \cite{huang2017densely}, deeper layers in DenseNet tend to rely more on high-level features rather than low-level features, which motivated us to improve the network by strengthening short connections. We found this can be achieved by gradually increasing the growth rate with depth. This increases the proportion of features from later layers relative to those from earlier layers.

For simplicity, we set the growth rate as $k=2^{m-1}k_0$, where $m$ is the dense block index and $k_0$ is a constant. This way of setting the growth rate does not introduce any additional hyperparameters. The "increasing growth rate" strategy places a larger proportion of parameters in the later layers of the model. This significantly improves computational efficiency, though it may reduce parameter efficiency in some cases. Depending on specific hardware constraints, trading one for the other may be advantageous \cite{liu2017learning}.

\begin{figure}[h]
\centering
\includegraphics[width=0.9\linewidth]{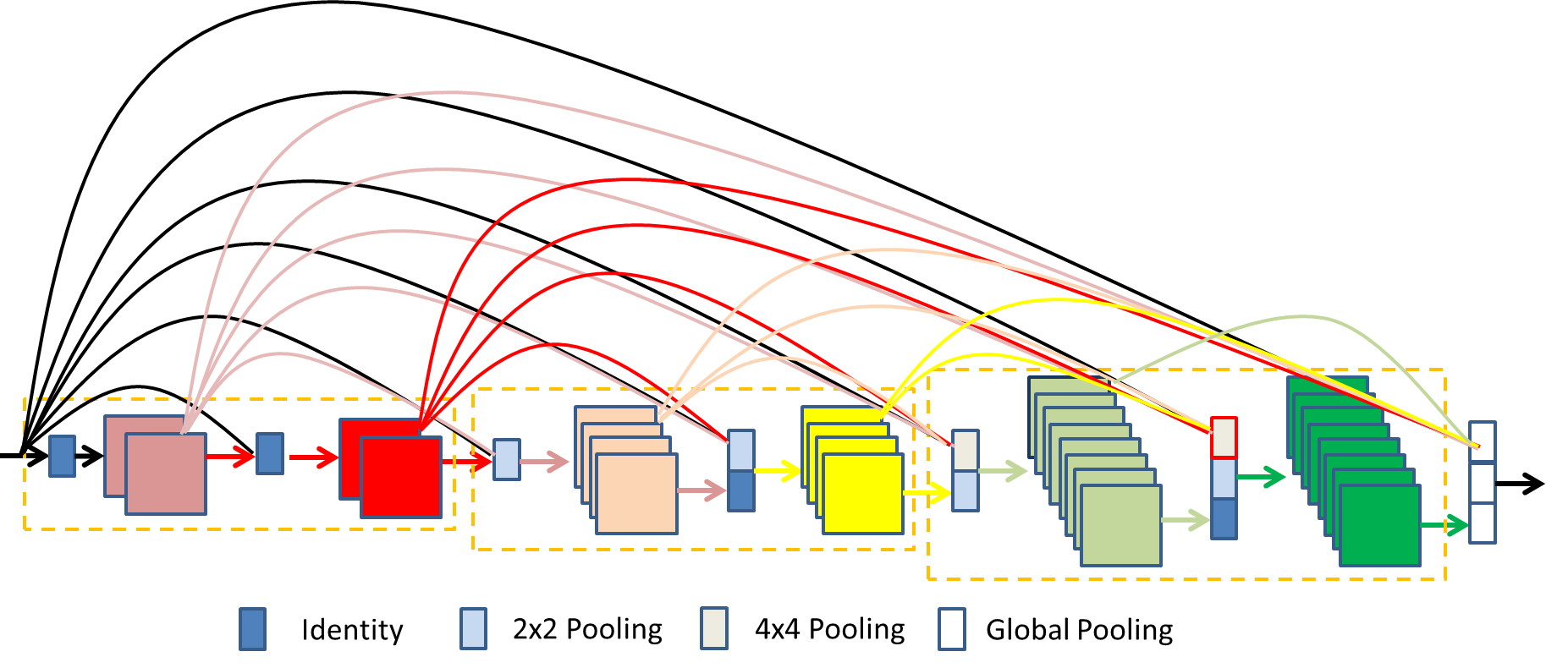}
\caption{The proposed DenseNet variant. It differs from the original DenseNet in two aspects: (1) layers with different resolution feature maps are also directly connected; (2) the growth rate doubles whenever the feature map size shrinks (features generated in the third yellow dense block are much more numerous than those generated in the first block).}
\label{fig:densenet_variant}
\end{figure}

\subsubsection{Fully Dense Connectivity}
To encourage more feature reuse than in the original DenseNet architecture, we connect the input layer to all subsequent layers in the network, even those in different dense blocks (see Figure \ref{fig:densenet_variant}). Since dense blocks have different feature resolutions, we downsample higher-resolution feature maps using average pooling when using them as input to lower-resolution layers.

The overall model architecture is shown below:

\begin{figure}[h]
\centering
\includegraphics[width=0.9\linewidth]{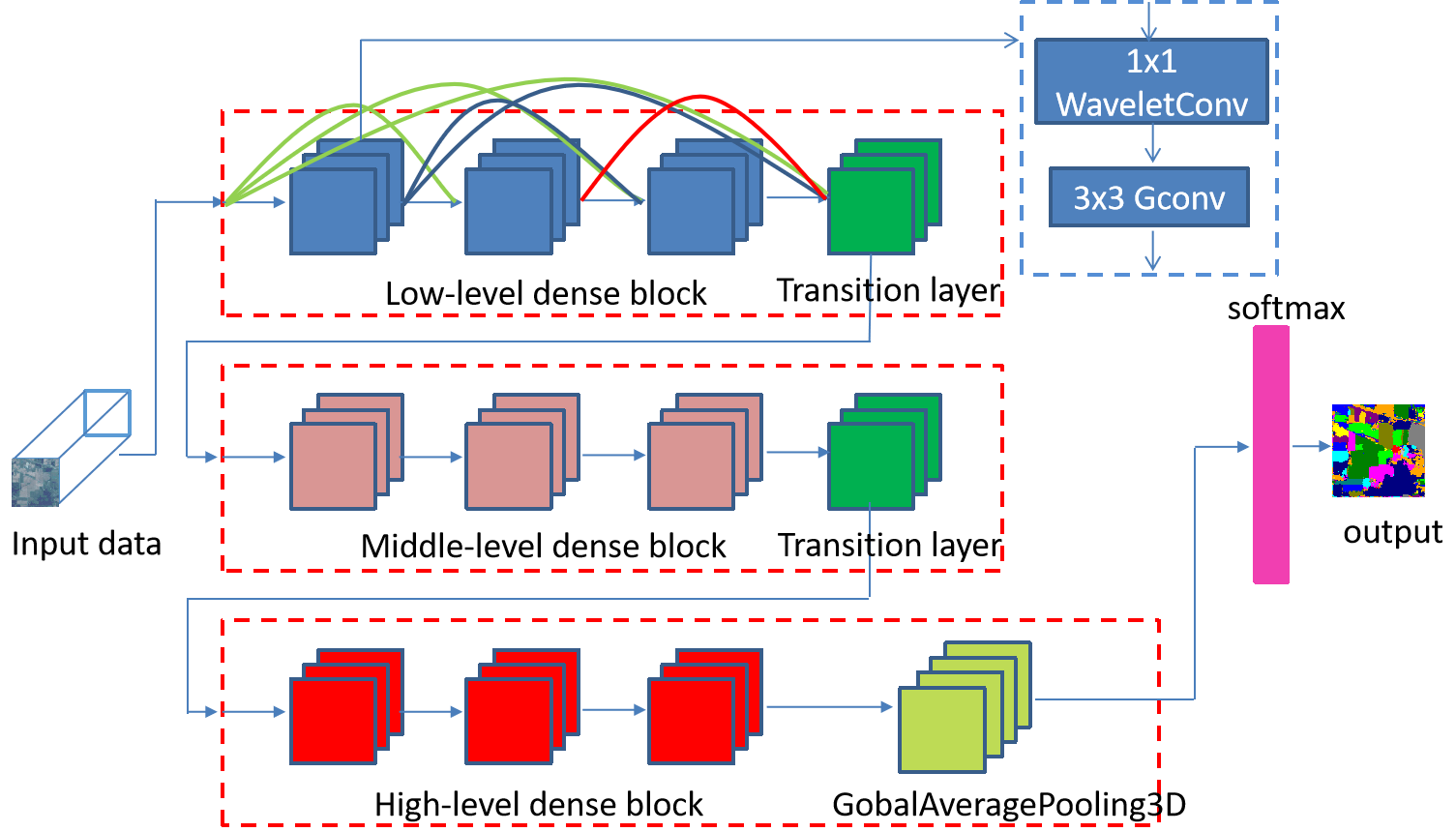}
\caption{The network structure of our WCNet, incorporating the basic architecture of 3D-DenseNet.}
\label{fig:network_structure}
\end{figure}

\section{Experiments and Analysis}
To evaluate the performance of the WCNet model, we conducted experiments on three representative hyperspectral image datasets: the Indian Pines test site image, the University of Pavia urban area image, and the Kennedy Space Center (KSC) dataset. The classification metrics include Overall Accuracy (OA), Average Accuracy (AA), and Kappa coefficient.

\subsection{Experimental Datasets}

\subsubsection{Indian Pines Dataset}
\begin{figure}[h]
\centering
\includegraphics[width=0.8\linewidth]{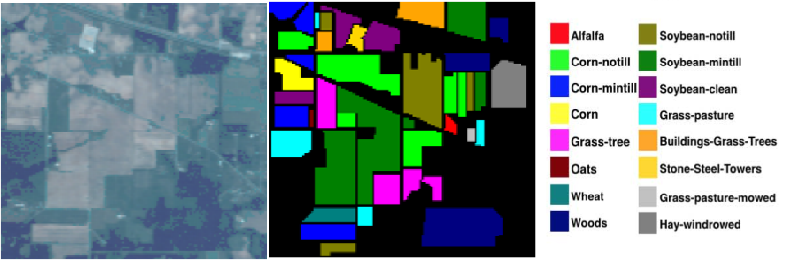}
\caption{False color image and ground-truth labels of Indian Pines dataset}
\label{fig:indian_pines}
\end{figure}

The Indian Pines dataset was collected in June 1992 by the AVIRIS sensor over the Indian Pines test site in northwestern Indiana, USA. The image size is $145 \times 145$ pixels with a spatial resolution of 20\,m, containing 220 spectral bands in the wavelength range of $0.4$--$2.5\,\mu\text{m}$. In our experiments, we removed 20 bands affected by water absorption and low signal-to-noise ratio (SNR), retaining 200 bands for analysis. The dataset contains 16 land-cover classes including grasslands, buildings, and crops. Figure~\ref{fig:indian_pines} shows the false-color image and ground-truth labels, while Table~\ref{tab:indian_samples} presents the sample distribution for each class.

\begin{figure}[h]
\centering
\includegraphics[width=0.8\linewidth]{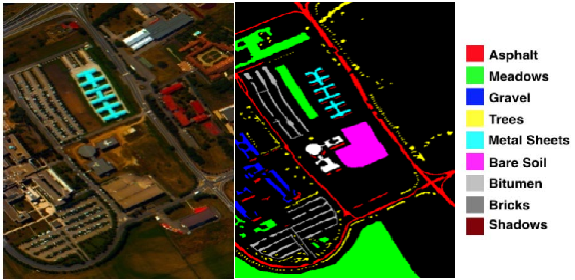}
\caption{False color image and ground-truth labels of Pavia University dataset}
\label{fig:pavia}
\end{figure}

\subsubsection{Pavia University Dataset}
The Pavia University dataset was acquired in 2001 by the ROSIS sensor over the University of Pavia, Italy. The image size is $610 \times 340$ pixels with a spatial resolution of 1.3\,m, containing 115 spectral bands in the wavelength range of $0.43$--$0.86\,\mu\text{m}$. We removed 12 noisy and water-absorption bands, retaining 103 bands for experiments. The dataset contains 9 land-cover classes including roads, trees, and roofs. Figure~\ref{fig:pavia} shows the spatial distribution of different classes, and Table~\ref{tab:pavia_samples} presents the sample allocation.

\subsubsection{Kennedy Space Center (KSC) Dataset}
The KSC dataset was collected on March 23, 1996 by the AVIRIS sensor over the Kennedy Space Center, Florida. AVIRIS acquired 224 bands with 10 nm width, centered at 400-2500 nm wavelengths. The data was acquired from an altitude of about 20 km with a spatial resolution of 18 m. After removing water absorption and low SNR bands, 176 bands were used for analysis, defining 13 land-cover classes.

\begin{table*}[!ht]
\begin{minipage}{\textwidth}
\centering
\makeatletter
\def\@makecaption#1#2{%
    \vskip\abovecaptionskip
    \centering 
    \small #1: #2\par
    \vskip\belowcaptionskip
}
\makeatother
\caption{OA, AA and Kappa metrics for different training set ratios on the Indian Pines dataset}
\begin{adjustbox}{width=0.5\columnwidth}
\begin{tabular}{cccc}
\toprule
Training Ratio & OA & AA & Kappa \\
\midrule
2:1:7 & 91.85 & 90.25 & 90.71 \\
3:1:6 & 96.57 & 93.30 & 96.09 \\
4:1:5 & 98.81 & 98.84 & 98.64 \\
5:1:4 & 99.22 & 98.37 & 99.11 \\
6:1:3 & 99.71 & 99.61 & 99.67 \\
\bottomrule
\end{tabular}
\label{tab:indian_ratios}
\end{adjustbox}
    \end{minipage}

 \vspace*{10pt} 
 
\begin{minipage}{\textwidth}
\centering
\makeatletter
\def\@makecaption#1#2{%
    \vskip\abovecaptionskip
    \centering 
    \small #1: #2\par
    \vskip\belowcaptionskip
}
\makeatother
\caption{OA, AA and Kappa metrics for different training set ratios on the Pavia University dataset}
\begin{tabular}{cccc}
\toprule
Training Ratio & OA & AA & Kappa \\
\midrule
2:1:7 & 99.01 & 98.58 & 98.69 \\
3:1:6 & 99.83 & 99.76 & 99.77 \\
4:1:5 & 99.93 & 99.91 & 99.91 \\
5:1:4 & 99.96 & 99.96 & 99.95 \\
6:1:3 & 99.98 & 99.97 & 99.98 \\
\bottomrule
\end{tabular}
\label{tab:pavia_ratios}
        \end{minipage}

 \vspace*{10pt} 
 
\begin{minipage}{\textwidth}        
\centering
\makeatletter
\def\@makecaption#1#2{%
    \vskip\abovecaptionskip
    \centering 
    \small #1: #2\par
    \vskip\belowcaptionskip
}
\makeatother
\caption{OA, AA and Kappa metrics for different training set ratios on the KSC dataset}
\begin{tabular}{cccc}
\toprule
Training Ratio & OA & AA & Kappa \\
\midrule
2:1:7 & 97.17 & 96.27 & 96.85 \\
3:1:6 & 98.21 & 97.71 & 98.00 \\
4:1:5 & 99.35 & 99.16 & 99.27 \\
5:1:4 & 99.52 & 99.30 & 99.47 \\
6:1:3 & 99.68 & 99.59 & 99.64 \\
\bottomrule
\end{tabular}
\label{tab:ksc_ratios}
        \end{minipage}
\end{table*}

\subsection{Experimental Analysis}
WCNet was trained for 80 epochs on all three datasets using the Adam optimizer. The experiments were conducted on a platform with four 80GB A100 GPUs. The network architecture employs a 3-stage structure with 4, 6, and 8 dense blocks in each stage respectively, growth rates of 8, 16, and 32, 4 heads, 3$\times$3 group convolution with 4 groups, gate\_factor=0.25, and compression=16.

\begin{table*}[!ht]
\begin{minipage}{\textwidth}
\centering
\makeatletter
\def\@makecaption#1#2{%
    \vskip\abovecaptionskip
    \centering 
    \small #1: #2\par
    \vskip\belowcaptionskip
}
\makeatother
\caption{OA, AA and Kappa metrics for different block sizes on Indian Pines}
\begin{tabular}{cccc}
\toprule
Block Size (M=N) & OA & AA & Kappa \\
\midrule
7 & 98.24 & 97.46 & 97.99 \\
9 & 99.32 & 99.34 & 99.22 \\
11 & 99.71 & 99.61 & 99.67 \\
13 & 99.58 & 98.72 & 99.52 \\
15 & 99.87 & 99.75 & 99.85 \\
17 & 99.87 & 98.98 & 99.85 \\
\bottomrule
\end{tabular}
\label{tab:indian_blocks}
        \end{minipage}

 \vspace*{10pt} 
 
\begin{minipage}{\textwidth}
\centering
\makeatletter
\def\@makecaption#1#2{%
    \vskip\abovecaptionskip
    \centering 
    \small #1: #2\par
    \vskip\belowcaptionskip
}
\makeatother
\caption{OA, AA and Kappa metrics for different block sizes on Pavia University}
\begin{tabular}{cccc}
\toprule
Block Size (M=N) & OA & AA & Kappa \\
\midrule
7 & 99.92 & 99.92 & 99.90 \\
9 & 99.98 & 99.96 & 99.97 \\
11 & 99.98 & 99.97 & 99.98 \\
13 & 99.99 & 99.99 & 99.99 \\
15 & 1 & 1 & 1 \\
17 & 1 & 1 & 1 \\
\bottomrule
\end{tabular}
\label{tab:pavia_blocks}
        \end{minipage}

 \vspace*{10pt} 
 
\begin{minipage}{\textwidth}
\centering
\makeatletter
\def\@makecaption#1#2{%
    \vskip\abovecaptionskip
    \centering 
    \small #1: #2\par
    \vskip\belowcaptionskip
}
\makeatother
\caption{OA, AA and Kappa metrics for different block sizes on KSC dataset}
\begin{tabular}{cccc}
\toprule
Block Size (M=N) & OA & AA & Kappa \\
\midrule
7 & 98.46 & 97.51 & 98.28 \\
9 & 98.72 & 98.40 & 98.57 \\
11 & 99.49 & 99.27 & 99.43 \\
13 & 99.87 & 99.84 & 99.86 \\
15 & 1 & 1 & 1 \\
17 & 99.87 & 99.81 & 99.86 \\
\bottomrule
\end{tabular}
\label{tab:ksc_blocks}
        \end{minipage}
\end{table*}

\subsubsection{Data Partitioning Ratio}
For hyperspectral data with limited sample sizes, the proportion of the training set significantly affects the stability and generalization ability of model performance. To thoroughly evaluate the sensitivity of data partitioning strategies, this study conducted a comparative analysis of model performance under different ratios of training, validation, and test sets. Experimental results show that when training samples are limited, adopting a 6:1:3 split ratio effectively balances the model's learning capacity and evaluation reliability---this configuration allocates 60\% of the samples for sufficient training, utilizes a 10\% validation set to implement an early stopping mechanism to prevent overfitting, and reserves 30\% for the test set to ensure statistically significant evaluation. WCNet adopted this 6:1:3 split ratio across the Indian Pines, Pavia University, and KSC benchmark datasets, with an 11$\times$11 neighboring pixel block size to balance local feature extraction and the integrity of spatial contextual information.

\subsubsection{Neighboring Pixel Blocks}
The network performs edge padding on the input image of size 145$\times$145$\times$103, transforming it into a 155$\times$155$\times$103 image (taking Indian Pines as an example). On this 155$\times$155$\times$103 image, an adjacent pixel block of size M$\times$N$\times$L is sequentially selected, where M$\times$N represents the spatial sampling size, and L denotes the full-dimensional spectrum. The original image is too large, which is not conducive to sufficient feature extraction through convolution, slows down the computation speed, increases short-term memory usage, and places high demands on the hardware platform. Therefore, processing with adjacent pixel blocks is adopted, where the size of the adjacent pixel block is an important hyperparameter. However, the range of the adjacent pixel block cannot be too small, as it may lead to an insufficient receptive field for the convolution kernel's feature extraction, resulting in poor local performance. As shown in Tables \ref{tab:indian_blocks}-\ref{tab:ksc_blocks}, on the Indian Pines dataset, as the pixel block size increases from 7 to 17, there is a noticeable improvement in accuracy. This trend is also evident in the Pavia University dataset. However, as the pixel block range continues to expand, the overall accuracy growth diminishes, exhibiting a clear threshold effect, which is similarly observed in the KSC dataset. Therefore, we selected a neighboring pixel block size of 15 for the Indian Pines dataset, 11 for the Pavia University dataset, and 17 for the KSC dataset.

\subsubsection{Network Parameters}
We divided WCNet into two types: base and large. The Table \ref{tab:wcnet_config} below presents the OA, AA, and Kappa results tested on the Indian Pines dataset. In the large model, the number of dense blocks in each of the three stages is set to 14, resulting in a significantly larger parameter count compared to the base model. However, the accuracy does not exhibit a substantial proportional increase relative to our base version, showing a clear effect of diminishing marginal returns. For both the base and large versions used in the comparison, the training set ratio is 6:1:3, and the adjacent pixel block size is 11.

\begin{table*}[!ht]
\centering
\makeatletter
\def\@makecaption#1#2{%
    \vskip\abovecaptionskip
    \centering 
    \small #1: #2\par
    \vskip\belowcaptionskip
}
\makeatother
\caption{WCNet model configurations and performance metrics}
\label{tab:wcnet_config}
\begin{tabular}{lccccc}
\toprule
Model & Stages/Blocks & Growth Rate & OA & AA & Kappa \\
\midrule
WCNet-base & 4,6,8 & 8,16,32 & 99.87 & 99.75 & 99.85 \\
WCNet-large & 14,14,14 & 8,16,32 & 99.93 & 99.77 & 99.93 \\
\bottomrule
\end{tabular}
\end{table*}

Using the thop library, we tested the parameter count (params) and computational complexity (GFLOPs) on the Indian Pines dataset (with spectral dimension of 200). As shown in Table \ref{tab:params_comparison}, our WCNet-base has extremely small parameter count while maintaining competitive model accuracy.

\begin{table*}[!ht]
\centering
\makeatletter
\def\@makecaption#1#2{%
    \vskip\abovecaptionskip
    \centering 
    \small #1: #2\par
    \vskip\belowcaptionskip
}
\makeatother
\caption{Comparison of parameters and computations of different models on the Indian Pines dataset}
\label{tab:params_comparison}
\begin{tabular}{lcccccc}
\toprule
 & 3D-CNN & 3D-DenseNet & HybridSN & LGCNet & DACNet & WCNet-base \\
\midrule
Params & 16,394,652 & 2,562,452 & 5,503,108 & 1,834,848 & 1,928,804 & 1,418,396 \\
FLOPs & 81,959,692 & 5,234,500 & 11,005,179 & 72,421,584 & 76,120,496 & 68,946,736 \\
\bottomrule
\end{tabular}
\end{table*}

\subsection{Experimental Results and Analysis}
\begin{figure*}[!ht]
\centering
\includegraphics[width=0.8\linewidth]{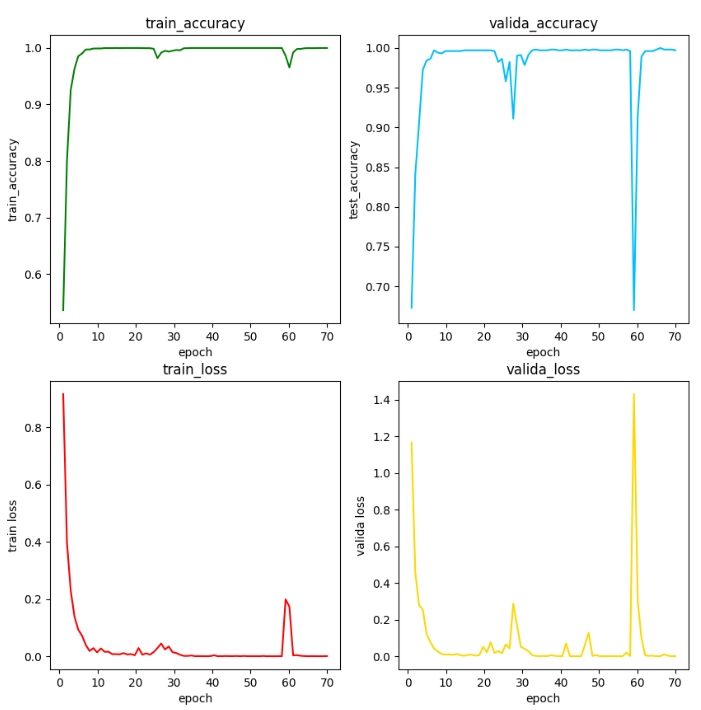}
\caption{Training and valid curves of WCNet showing loss and accuracy evolution}
\label{fig:training_curves}
\end{figure*}

On the Indian Pines dataset, the input size for WCNet is $17\times17\times200$; on the Pavia University dataset, it is $17\times17\times103$; and on the KSC dataset, it is $17\times17\times176$. The experiment compared SSRN, 3D-CNN, 3D-SE-DenseNet, Spectralformer, LGCNet, DGCNet, and our WCNet-base. As shown in Tables \ref{tab:indian_results}-\ref{tab:pavia_results}, WCNet achieved leading accuracy overall on both datasets. Fig.\ref{fig:training_curves} illustrates the loss and accuracy changes on the training and validation sets, demonstrating that the loss function converges very quickly and the accuracy increases in a stable manner.

\begin{table*}[t]
\centering
\caption{Classification accuracy comparison (\%) on Indian Pines dataset}
\label{tab:indian_results}
\resizebox{\textwidth}{!}{
\begin{tabular}{@{}lccccccc@{}}
\toprule
Class & SSRN & 3D-CNN & 3D-SE-DenseNet & Spectralformer & LGCNet & DGCNet & WCNet \\
\midrule
1 & 100 & 96.88 & 95.87 & 70.52 & 100 & 100 & 100 \\
2 & 99.85 & 98.02 & 98.82 & 81.89 & 99.92 & 99.47 & 100 \\
3 & 99.83 & 97.74 & 99.12 & 91.30 & 99.87 & 99.51 & 100 \\
4 & 100 & 96.89 & 94.83 & 95.53 & 100 & 97.65 & 100 \\
5 & 99.78 & 99.12 & 99.86 & 85.51 & 100 & 100 & 100 \\
6 & 99.81 & 99.41 & 99.33 & 99.32 & 99.56 & 99.88 & 100 \\
7 & 100 & 88.89 & 97.37 & 81.81 & 95.83 & 100 & 100 \\
8 & 100 & 100 & 100 & 75.48 & 100 & 100 & 100 \\
9 & 0 & 100 & 100 & 73.76 & 100 & 100 & 100 \\
10 & 100 & 100 & 99.48 & 98.77 & 99.78 & 98.85 & 100 \\
11 & 99.62 & 99.33 & 98.95 & 93.17 & 99.82 & 99.72 & 99.60 \\
12 & 99.17 & 97.67 & 95.75 & 78.48 & 100 & 99.56 & 100 \\
13 & 100 & 99.64 & 99.28 & 100 & 100 & 100 & 100 \\
14 & 98.87 & 99.65 & 99.55 & 79.49 & 100 & 99.87 & 100 \\
15 & 100 & 96.34 & 98.70 & 100 & 100 & 100 & 100 \\
16 & 98.51 & 97.92 & 96.51 & 100 & 97.73 & 98.30 & 96.43 \\
\midrule
OA & 99.62$\pm$0.00 & 98.23$\pm$0.12 & 98.84$\pm$0.18 & 81.76 & 99.85$\pm$0.04 & 99.58 & 99.87 \\
AA & 93.46$\pm$0.50 & 98.80$\pm$0.11 & 98.42$\pm$0.56 & 87.81 & 99.53$\pm$0.23 & 99.55 & 99.75 \\
K & 99.57$\pm$0.00 & 97.96$\pm$0.53 & 98.60$\pm$0.16 & 79.19 & 99.83$\pm$0.05 & 99.53 & 99.85 \\
\bottomrule
\end{tabular}
}
\end{table*}

\begin{figure}[t]
\centering
\includegraphics[width=0.9\linewidth]{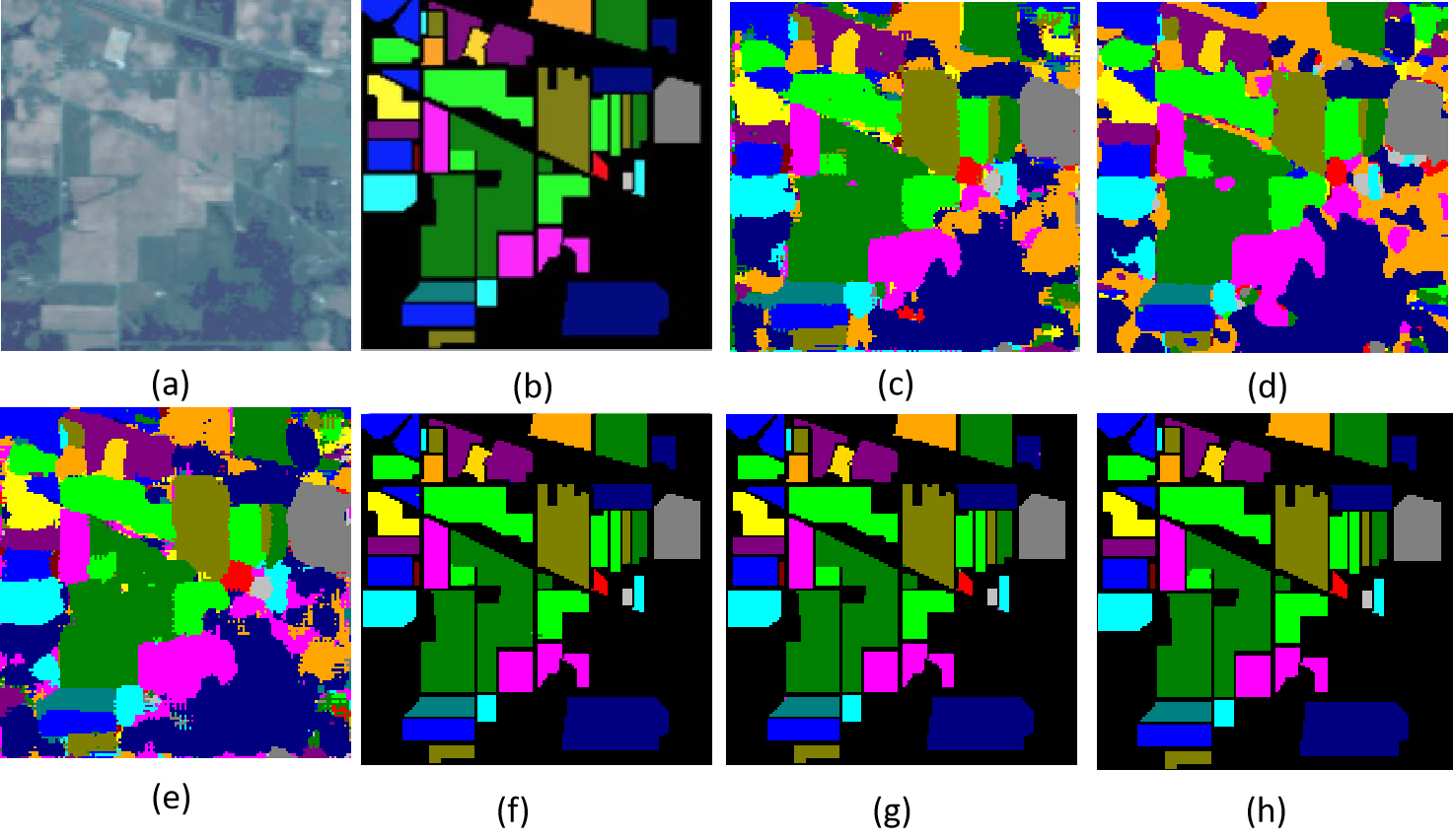}
\caption{Classification results on Indian Pines dataset: (a) Sample image, (b) Ground truth labels, (c)-(i) Results from SSRN, 3D-CNN, 3D-SE-DenseNet, LGCNet, DGCNet and WCNet}
\label{fig:indian_results}
\end{figure}

\begin{table*}[!ht]
\centering
\caption{Classification accuracy comparison (\%) on Pavia University dataset}
\label{tab:pavia_results}
\resizebox{\textwidth}{!}{
\begin{tabular}{@{}lcccccc@{}}
\toprule
Class & SSRN & 3D-CNN & 3D-SE-DenseNet & Spectralformer & LGCNet & WCNet \\
\midrule
1 & 89.93 & 99.96 & 99.32 & 82.73 & 100 & 100 \\
2 & 86.48 & 99.99 & 99.87 & 94.03 & 100 & 100 \\
3 & 99.95 & 99.64 & 96.76 & 73.66 & 99.88 & 100 \\
4 & 95.78 & 99.83 & 99.23 & 93.75 & 100 & 100 \\
5 & 97.69 & 99.81 & 99.64 & 99.28 & 100 & 100 \\
6 & 95.44 & 99.98 & 99.80 & 90.75 & 100 & 100 \\
7 & 84.40 & 97.97 & 99.47 & 87.56 & 100 & 100 \\
8 & 100 & 99.56 & 99.32 & 95.81 & 100 & 100 \\
9 & 87.24 & 100 & 100 & 94.21 & 100 & 100 \\
\midrule
OA & 92.99$\pm$0.39 & 99.79$\pm$0.01 & 99.48$\pm$0.02 & 91.07 & 99.99$\pm$0.00 & 100 \\
AA & 87.21$\pm$0.25 & 99.75$\pm$0.15 & 99.16$\pm$0.37 & 90.20 & 99.99$\pm$0.01 & 100 \\
K & 90.58$\pm$0.18 & 99.87$\pm$0.27 & 99.31$\pm$0.03 & 88.05 & 99.99$\pm$0.00 & 100 \\
\bottomrule
\end{tabular}
}
\end{table*}

\section{Conclusion}
This paper introduces Wavelet Convolution (Wavelet Conv) on 3D convolution kernels. This module employs cascaded wavelet transform (WT) decomposition and performs a series of small-kernel convolutions, where each convolution focuses on different frequency bands of the input signal with gradually increasing receptive fields. This process enables us to place greater emphasis on low-frequency components in the input while adding only a small number of trainable parameters. In fact, for a k×k receptive field, our trainable parameters grow logarithmically with k. To address the problems of excessive redundant information, increased computational load in feature extraction and fusion within 3D convolution-Transformer architectures, as well as the over-parameterization that leads to overfitting and reduced generalization capability, Wavelet Conv enhances the model's representation of joint spatial-spectral information by extending convolution kernels through wavelet transformation. This dynamic design enables WCNet to flexibly adapt to the feature requirements of different spatial regions without relying on a single static convolution kernel, while effectively skipping redundant information and reducing computational complexity.WCNet further leverages the 3D-DenseNet architecture to extract spatial structures and spectral information of more critical features, providing an efficient solution for hyperspectral image classification. It successfully addresses the challenges posed by sparse object distribution and spectral redundancy.

{\small
\bibliographystyle{template}
\bibliography{template}
}

\end{document}